\documentclass[runningheads]{llncs}
\usepackage{graphicx}
\usepackage{comment}
\usepackage{amsmath,amssymb} 
\usepackage{color}
\usepackage{subfigure}
\usepackage{array}
\usepackage{tabularx}
\usepackage{float}
\usepackage{epsfig}
\usepackage{multirow}
\usepackage[width=122mm,left=12mm,paperwidth=146mm,height=193mm,top=12mm,paperheight=217mm]{geometry}

\begin{document}

\pagestyle{headings}
\mainmatter
\title{Affinity-aware Compression and Expansion Network for Human Parsing}  \author{{Xinyan Zhang, Yunfeng Wang, and Pengfei Xiong}}
\institute{\{zhangxinyan, wangyunfeng, xiongpengfei\}@megvii.com}
\maketitle

\begin{abstract}
As a fine-grained segmentation task, human parsing is still faced with two challenges: inter-part indistinction and intra-part inconsistency, due to the ambiguous definitions and confusing relationships between similar human parts. To tackle these two problems, this paper proposes a novel \textit{Affinity-aware Compression and Expansion} Network (ACENet), which mainly consists of two modules: Local Compression Module (LCM) and Global Expansion Module (GEM).
Specifically, LCM compresses parts-correlation information through structural skeleton points, obtained from an extra skeleton branch. It can decrease the inter-part interference, and strengthen structural relationships between ambiguous parts. 
Furthermore, GEM expands semantic information of each part into a complete piece by incorporating the spatial affinity with boundary guidance, which can effectively enhance the semantic consistency of intra-part as well.
ACENet achieves new state-of-the-art performance on the challenging LIP and Pascal-Person-Part datasets. In particular, {58.1\%} mean IoU is achieved on the LIP benchmark.

\end{abstract}

\section{Introduction}

\begin{figure}[t]
	\begin{center}
		\begin{tabular}{cccc}
			\hspace{-1.5ex}
			\vspace{-1.6ex}
			\subfigure{
				\begin{tabular}{r}
					\includegraphics[width=.22\textwidth, height= 0.9in]{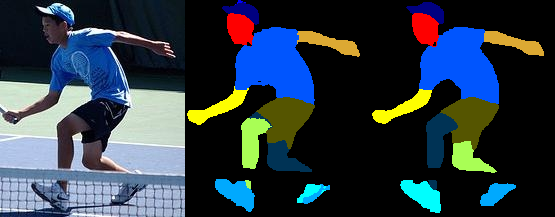}
				\end{tabular}		
			}  &
			\subfigure{
				\begin{tabular}{c}			
					\includegraphics[width=.22\textwidth, height= 0.9in]{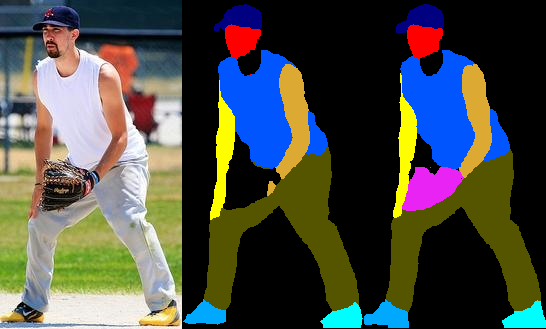}
				\end{tabular}
			} &
		    \subfigure{
		    	\begin{tabular}{c}					
		    		\includegraphics[width=.22\textwidth, height= 0.9in]{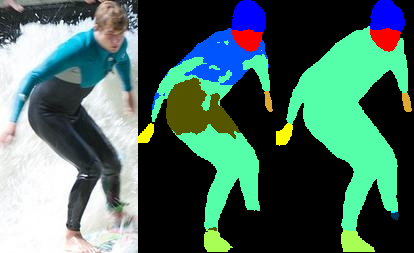}
		    	\end{tabular}
		    } &
	        \subfigure{
	        	\begin{tabular}{c}					
	        		\includegraphics[width=.22\textwidth, height= 0.9in]{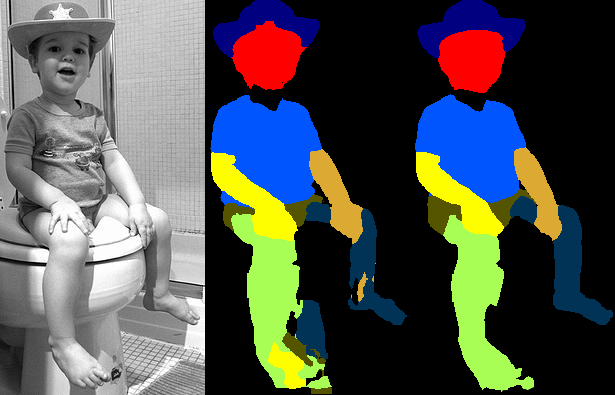}
	        	\end{tabular}
	        } \\
        	\vspace{-0.8ex}
			\hspace{-1ex}{\scriptsize (a)} & {\scriptsize (b)} & {\scriptsize (c)} & {\scriptsize (d)}\\
		\end{tabular}
		\vspace{-2.5ex}
		\includegraphics[width=.65\textwidth, height=0.18in]{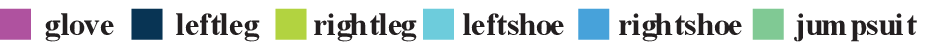}
	\end{center}
	\vspace{-3mm}
	\caption{Typical parsing network suffers from ``intra-part inconsistency" and ``inter-part indistinction". In each subfigure, the sequence of images is input, baseline, and ACENet result. In (a), baseline cannot recognize left/right shoes; in (b), the glove is missed; in (c), textures on jumpsuit are confused; in (d), legs under the similar background cannot be discriminated. ACENet effectively addresses the above issues via LCM and GEM.}
	\vspace{-1mm}
	\label{fig:intro}
\end{figure}

Human parsing, also known as human parts/clothes segmentation, aims at segmenting a human image into different semantic parts. It has many practical applications in industry, such as video surveillance, autonomous driving, human fashion, and virtual reality products. With the recent development of deep learning, many methods~\cite{chen2014semantic,chen2016attention,gong2017look,liang2016semantic2,liang2016semantic,ruan2019devil,sun2019high,xia2017joint} have obtained promising results on the benchmarks by using deep convolutional neural networks. However, most existing methods still regard human parsing as a general semantic segmentation problem, and they often improve the network performance by increasing receptive fields, or applying extra edge supervision.

Human parsing is a fine-grained segmentation task, which needs to accurately segment the whole human body into defined parts, e.g. arm, leg, jumpsuit. Different from the traditional semantic segmentation tasks, the definitions and relationships of human parts are often ambiguous and confused. As shown in Fig.~\ref{fig:intro}, complex postures and object occlusion may lead to wrong results, such as left-right confusion and missing parts. Meanwhile, the similar appearance between different semantic parts, such as dress and skirt, can also bring about the confusion and inconsistency under the same semantic label. Therefore, the previous human parsing methods are always struggling with ``inter-part indistinction" and ``intra-part inconsistency".

To address these two challenges, we rethink the human parsing task from new perspectives. 
On the one hand, different human parts have predefined structural relationships, while the distribution of various parts represents the information of body structure. We can thus compress them into correlative points to improve inter-part structure correlations and reduce the impact of complex conditions.
On the other hand, since each human part has its independent semantic information, we can expand each semantic part into a complete piece to enhance the intra-part consistency, as well as avoid the effect of confusion definitions.  
To this end, we propose an \textit{Affinity-aware Compression and Expansion Network} (ACENet) to concurrently solve these two problems.

\begin{figure}[t]
	\begin{center}
		\includegraphics[width=0.98\textwidth, height=1.85in]{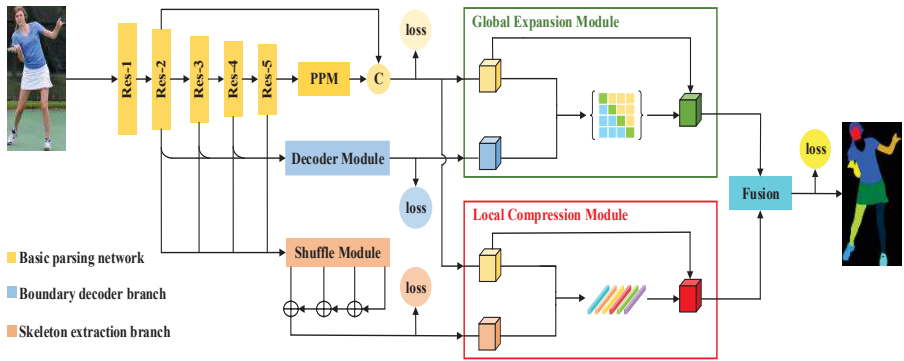}
	\end{center}
	\vspace{-5mm}
	\caption{The overall architecture of our model. An input image is fed to a ResNet-101 backbone to extract features. The output of \textit{Res-5} is designed as the input of a Pyramid Pooling Module (PPM) to obtain multi-scale information. Then it concatenates with the output of \textit{Res-2} to get basic parsing features (yellow block), following the scheme in CE2P~\cite{ruan2019devil}. In addition, features from the backbone are fed to a decoder module for constructing boundary features (blue block). Skeleton features (orange block) are obtained by a channel shuffle module. With the parsing and skeleton features as inputs, the Local Compression Module enhances discriminative ability with skeleton supervision. With parsing and boundary features as inputs, the Global Expansion Module further refines the parsing results with the boundary guidance. At last, the output features of both modules are concatenated to predict the final parsing result.}
	\vspace{-2mm}
	\label{fig:overall structure}
\end{figure}

Specifically, as illustrated in Fig.~\ref{fig:overall structure}, the proposed ACENet is mainly comprised of two modules: Local Compression Module (LCM) and Global Expansion Module (GEM). The LCM is designed to address the ``inter-part indistinction" issue. To improve the distinguish ability of the network, we adopt a structure point branch to extract the structural relationships of the human body. By combing feature-maps from the point branch with parsing features, we construct an affinity matrix with the channel manner, and each row in the affinity matrix represents the aggregation power of a structural point to segmentation parts. In addition, these structure points can be defined as skeleton points or the centers of human parts. Based on the explicit constraint, the LCM further adopts the affinity attention map onto the coarse parsing result to correct the indistinguishable regions. As shown in Fig.~\ref{fig:intro}(a), the correct categories and location of all components are obtained, even with twisted posture.

Furthermore, GEM is built to model semantic relationships between adjacent patches, and merge patches with the same category into a complete one. 
Based on the boundary supervision, GEM generates an affinity matrix in the spatial manner to strengthen the internal semantic consistency of each part. 
The spatial-affinity map between boundary and parsing features can also alleviate the impact of similar textures between different parts.
The purpose of GEM is to explicitly represent the intra-part relationship and segment confusing parts with the affinity module, while some existing approaches~\cite{ruan2019devil} implicitly facilitated the result by feature fusion with edge supervision. 
As shown in Fig.~\ref{fig:intro}(c), the jumpsuit is repaired as a complete part with the help of GEM. 

The proposed LCM compresses the segmentation information into structural representations, and GEM expands the information into independent semantic descriptions. ACENet is the first attempt to introduce the affinity-aware modules on human parsing, rather than direct fusion of multi-feature. We validate the effectiveness of the proposed ACENet on two popular benchmarks, LIP\cite{gong2017look} and Pascal-Person-Part \cite{chen2014detect}. Compared with the state-of-the-art methods, ACENet achieves 58.1\% mIoU on LIP and 70.9\% mIoU on Pascal-Person-Part.

In summary, the main contributions of this work are listed as follows:
\begin{itemize}
	\item[$\bullet$] We propose a novel Affinity-aware Compression and Expansion Network (ACENet) to address the indistinction and inconsistency problems in human parsing. We further rethink the multi-part parsing by abstracting it into the aspects of structure and semantics, and effectively fuse them with their affinity relationships.
	\item[$\bullet$] We introduce the Local Compression Module (LCM) to enhance the discriminative ability with skeleton guidance, as well as the Global Expansion Module (GEM) to improve the consistency of intra-part through the spatial affinity with boundary guidance.
	\item[$\bullet$] Our method achieves the state-of-the-art performance on two popular human parsing datasets (LIP and Pascal-Person-Part), especially on the large scale benchmark LIP.
\end{itemize}

\section{Related Work}
\noindent\textbf{Human Parsing.} Recently human parsing has achieved huge progress with the strong learning power of deep neural networks and largely available annotated datasets. Some previous works \cite{liang2018look,xia2017joint} solved human parsing and pose estimation in a multi-task framework, based on the observation that there is a strong correlation between these two tasks. Besides, edge information is exploited to obtain sharp and semantic reasonable parsing results \cite{ruan2019devil}. In \cite{gong2018instance}, human parts are explicitly learned by a Part Grouping Network, followed by an edge detection to group parts to instances. In \cite{liang2016semantic2,liang2016semantic}, a Graph-based LSTM \cite{hochreiter1997long} is proposed to construct an undirected graph using superpixels as nodes and the spatial relations of the superpixels as edges. In \cite{gong2019graphonomy}, a human parsing agent, which incorporates hierarchical graph transfer learning on a parsing network to encode the underlying label semantic structures and propagate relevant semantic information is proposed to parse image from different domains and different granularity. In \cite{wang2019learning}, parsing human parts was solved in a hierarchical form, in which the information from the direct, top-down, and bottom-up inference processes is combined to get final parsing results. Also, representations of high resolution and object-context were proved to be useful for human parsing \cite{sun2019high,yuan2019object}. 

\noindent\textbf{Attention.} The attention mechanism is widely used in computer vision and natural language processing communities \cite{sutskever2014sequence,vaswani2017attention}. Recently, self-attention is proved to be a vital module in semantic segmentation. In \cite{wang2018non}, a Non-local module is proposed to capture long-range spatial-temporal dependencies in a single image or sequence of images. A series of works inherit the thoughts in this paper and developed many methods to use self-attention to solve semantic segmentation tasks \cite{fu2019dual,yuan2019object,zhang2019acfnet}. In order to reduce computational complexity of self-attention matrix, SVD \cite{golub1971singular} and low rank factorization are used \cite{huang2019ccnet,yue2018compact,zhang2019latentgnn,zhao2018psanet}. Different from existing methods, we use information from different domains to construct our attention blocks. 

\noindent\textbf{Multi-task Learning.} Due to the strong correlations between human pose estimation and human parsing, there are plenty of methods jointly solve these two tasks. In \cite{xia2017joint}, Xia \emph{et al.} use two fully convolutional networks to estimate pose joint potential and semantic part potential firstly. Then, a Fully-connected Conditional Random Field (FCRF) \cite{chen2014semantic} is used to encourage semantic and spatial consistency between parts and joints. In \cite{gong2017look}, a part module and a joint module are appended to a feature extraction backbone to capture the part
context and keypoint context which are used to generate parsing and pose results, respectively. Finally, a refinement network is performed
based on the predicted maps and generated context to produce better results. In \cite{nie2018mutual}, Nie \emph{et al.} use an inner-network mutual adaptation module to build dynamic interaction and cooperation between parsing and pose estimation to exploit their mutual benefits.

\section{Methodology}
In this section, we first introduce the proposed Local Compression Module and Global Expansion Module in detail, and then elaborate on how these two modules improve human parsing results through the supervisions of skeleton and boundary features respectively. Finally, we describe the complete encoder-decoder network architecture, named Affinity-aware Compression and Expansion Network (ACENet). 

\begin{figure}[t]
	\begin{center}
		\begin{tabular}{cc}
			\vspace{-1.6ex}
			\subfigure{
				\begin{tabular}{l}
					\includegraphics[width=.46\textwidth, height= 1.35in]{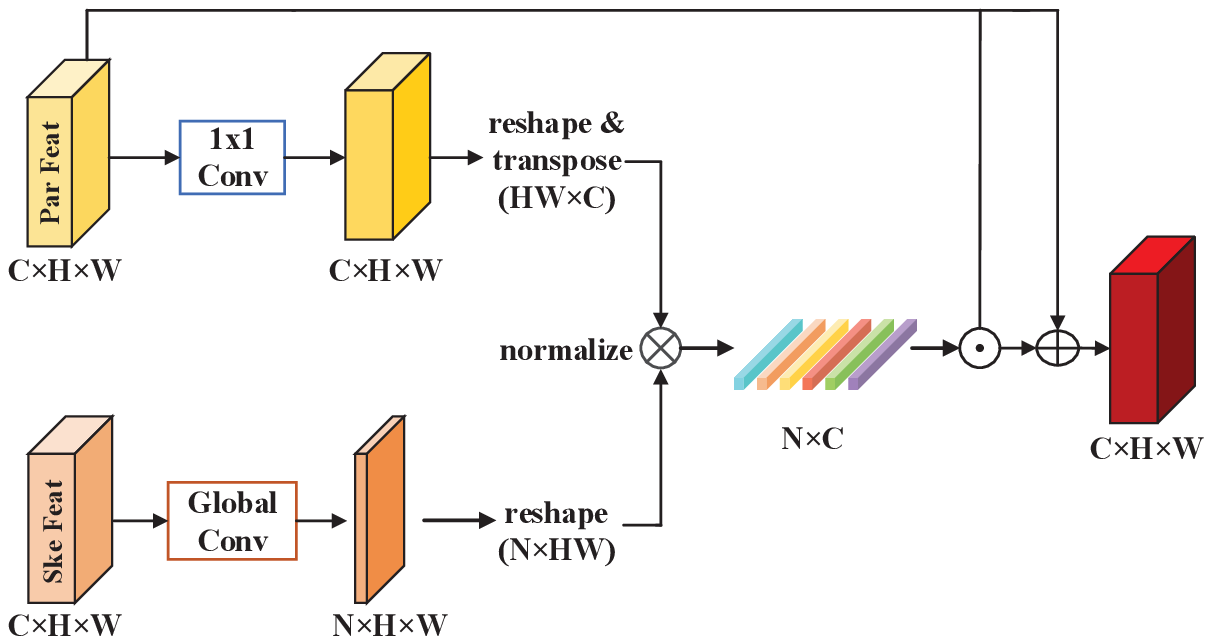}
				\end{tabular}		
			}  &
			\subfigure{
				\begin{tabular}{r}			
					\includegraphics[width=.46\textwidth, height= 1.35in]{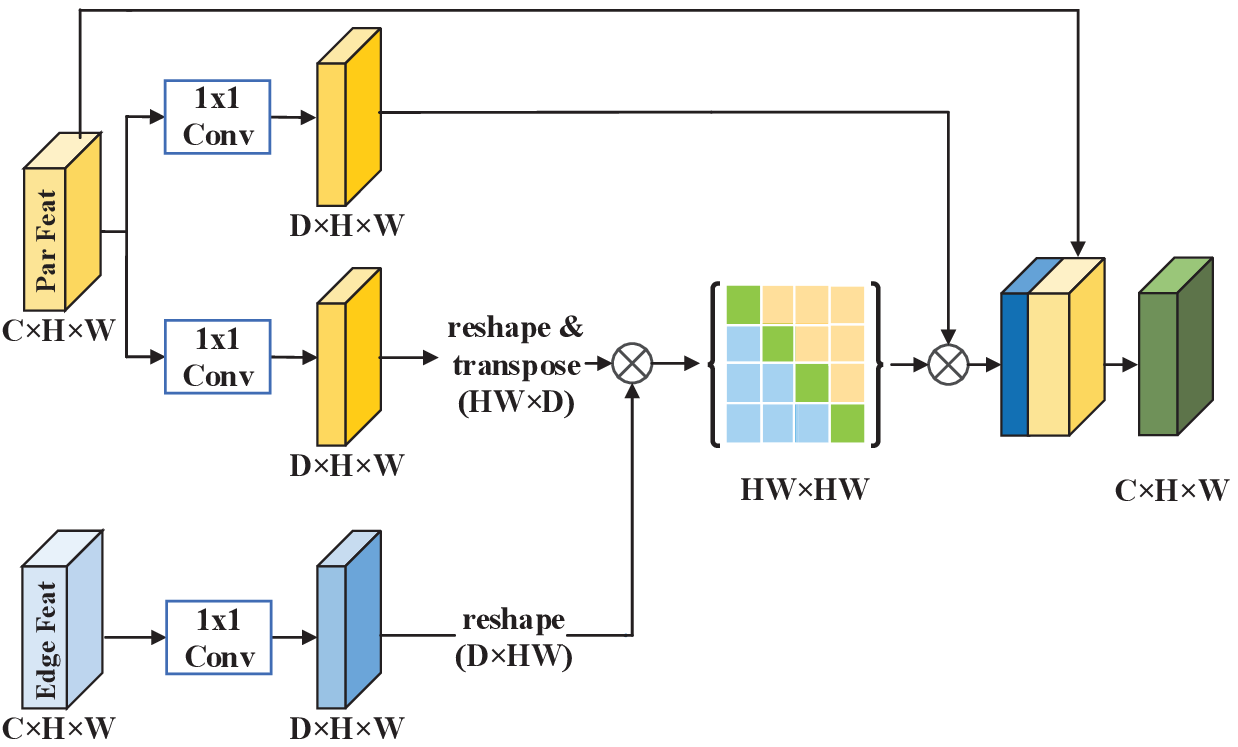}
				\end{tabular}
			} \\
			\vspace{-0.8ex}
			{\scriptsize (a) LCM} & {\scriptsize (b) GEM}\\
		\end{tabular}
		\vspace{-2ex}
	\end{center}
	\caption{(a) The details of Local Compression Module (LCM). (b) The details of t Global Expansion Module (GEM).}
	\label{fig:lcm&gem}
\end{figure}

\subsection{Local Compression Module}
\label{sec:lcm}

Human parsing divides persons into multiple semantic categories and shows the details of human-beings. However, in practice, the prediction of human parts can easily get wrong when people are in left-right mode or under complex conditions, which is so-called ``inter-part indistinction'' or ``inter-part confusion''. 

This problem is mainly caused by the lack of perception and distinction between correlation parts, such as left-right arms and shoes. To address this issue, we propose the Local Compression Module (LCM), which can build guidance for parsing features via skeletal representations, and exploit richer structural context information. Different from pixel-wise representation in the human parsing task, skeletons use the two-dimensional center points to represent the locations of different human body parts. Therefore, the skeleton can be regarded as compression of local segmentation information for human parts.

The structure of LCM is shown in Fig.~\ref{fig:lcm&gem}(a). Given a skeleton feature $S\in \mathbb{R}^{C\times H \times W}$, we first feed it into a Global Convolutional block (GC), which employs a combination of $1\times k + k\times 1$ and $k\times 1 + 1\times k$ convolutional layers. Intuitively, the GC block can encode the features with a large kernel manner to extract correlation information in skeleton features, and reduce the number of channels from $C$ to $N$. Then we reshape the encoded features to $M\in \mathbb{R}^{N\times HW}$. 

At the same time, the parsing feature $P$ is encoded by a $1\times1$ convolutional layer. We reshape and transpose $P$ to $\hat{P}\in\mathbb{R}^{HW\times C}$. After that, we construct the compression affinity matrix $A\in \mathbb{R}^{N\times C}$ by combing $M$ and $\hat{P}$ with a matrix multiplication, and $A$ is further normalized ranging in $[0, 1]$ for better representation ability, 
\begin{align}
\left. A_{ij}\, =\, {e^{M^i\cdot \hat{P}^j}}\, \middle/ \, {\sum\nolimits_{j=1}^{C}{e^{M^i\cdot \hat{P}^j}}} \right. ,\;
\end{align} 
where $\hat{P}^j$ denotes the parsing feature-map at $j$-th channel and $M^i$ denotes the encoded correlation map at $i$-th channel. Specifically, each values in 2D maps of the encoded features $M$ indicates the skeleton-guided weights for each corresponding spatial pixels in parsing features $\hat{P}$. Each row in $A$ represents the aggregation power of 2D skeleton-guidance maps for parsing features. $A$ can be thus considered as compressed channel characteristics of the parsing feature $\hat{P}$ by guidance of the encoded skeleton features $M$.

Finally, in order to enhance the inter-part distinction ability, we should strengthen the discriminative features and inhibit the indiscriminative ones. Therefore, we adopt channel-wise multiplication between channel-wise affinity matrix $A=[a_1,\cdots,a_c,\cdots,a_C]$ and parsing features $P=[p_1,\cdots,p_c,\cdots,p_C]'$ as a feature attention module, since $A$ represents the channel relationships and feature selection of parsing features with skeleton guidance. Furthermore, we obtain final output $L\in \mathbb{R}^{C\times H \times W}$ with a element-wise sum operation to alleviate the gradient vanishing problem, and it can also avoid the information flow in LCM affecting the other parts of network,  
\begin{align}
l_{c}=\frac{1}{N}\sum_{i=1}^{N}F_{scale}(a_c^i,p_c)+p_c,\;
\end{align} 
where $L=[l_1,\cdots,l_c,\cdots,l_C]$ and $F_{scale}$ refers to channel-wise multiplication between the feature map $p_c\in \mathbb{R}^{H\times W}$ and the skeleton-guidance compression weights $a_c\in \mathbb{R}^N$. The dimension $N$ in $A$ controls the intensity of compression on parsing feature, which is further illustrated in Section~\ref{sec_lcm}.1.

\subsection{Global Expansion Module}
\label{sec:gem}

The boundary information, which is extracted from parsing labels, can be regarded as a representation of global information, expanding each parsing part to a reasonable boundary. We introduce the Global Expansion Module (GEM) to improve the network performance and increase the consistency of ambiguous parts with boundary supervision (as shown in Fig.~\ref{fig:lcm&gem}(b)). In the proposed GEM, we adopt a spatial affinity attention module between parsing and boundary features, instead of the traditional self-correlation of boundary information. It can effectively help to construct spatial affinity guidance for better prediction.

Given the parsing feature $P\in\mathbb{R}^{C\times H\times W}$ and the boundary feature $E\in\mathbb{R}^{C\times H\times W}$, we first use two $1\times 1$ convolutional layers to generate the corresponding encoded features $\{P_e, E_e\}$ to reduce the channel number from $C$ to $D$ for making the proposed GEM with less computational cost, and further reshape them to $\mathbb{R}^{D\times HW}$. While both parsing and boundary features are extracted from the same backbone, the dimensions of them are set equal for convenience. Then we calculate the spatial affinity matrix $G\in\mathbb{R}^{HW\times HW}$ with the transpose of $P_e$ and $E_e$ through matrix multiplication:
\begin{align}
\left. G_i\, =\, {e^{P_e'\cdot E_e^i}}\, \middle/ \, {\sum\nolimits_{j=1}^{HW}e^{P_e^j\cdot E_e^i}} \right. ,\;
\end{align}
where $G=[G_1,\cdots,G_i,\cdots,G_{HW}]$, and $G_i$ demonstrates spatial constraint relationships for parsing feature-maps from the $i$-th position of boundary. The greater the correlation, the pixels more likely belong to the same part. In this way, we can explicitly aggregate the global representation of parsing features by applying boundary supervision. 

Furthermore, we utilize a $1\times 1$ convolutional layer to obtain $K\in\mathbb{R}^{D\times H\times W}$ from parsing feature $P$, and reshape it to $D\times HW$. As a standard attention module, we combine $K$ and the spatial-affinity matrix $G$ with matrix multiplication ($R=K\cdot G$), and then reshape the guided output $R$ to $\mathbb{R}^{D\times H\times W}$. With boundary guidance, $R$ represents the enhanced semantic information of intra-part through a global contextual view, and it can thus improve intra-part consistency.   

At last, we apply a concatenation operation to combine $R$ and $P$, since it is helpful to incorporate original parsing features $P$ to provide an extra supervision,
\begin{align}
X=\delta([P,R]),\;
\end{align}
where $\delta$ is $1\times 1$ convolutional function for dimension reduction, $X\in\mathbb{R}^{C\times H \times W}$ is the final output of GEM. In addition, the impact of the hyper-parameter $D$ in GEM will be further discussed in section~\ref{sec_gem}.2.

\subsection{Network Architecture}
\label{sec: net_arch}

Our method integrates the proposed Local Compression Module and Global Expansion Module into a unified network. The overall architecture is shown in Fig.~\ref{fig:overall structure}. Specifically, our model consists of four components: basic parsing network, boundary decoder module, skeleton generation module and the enhanced attention group including LCM and GEM.

\textbf{Basic Parsing Network.} As for baseline network, we use a modified ResNet-101 \cite{he2016deep} as backbone, initialized with weights learned on ImageNet \cite{deng2009imagenet}. In detail, the dilated convolutional layer with a rate of 2 is applied to the last \textit{Res-5} block, so the output size is 1/16 of the resolution of the input image. The Pyramid Pooling Module (PPM) \cite{zhao2017pyramid} is also adopted to effectively obtain the multi-scale context information from the backbone network. Following the scheme in~\cite{ruan2019devil}, we concatenate the output feature of \textit{Res-5} and the context features with high-resolution details, which is captured from \textit{Res-2} stage. A $1\times 1$ convolutional layer is used to generate the basic parsing results of our model.

\textbf{Skeleton Extraction Branch.} This module generates the pose skeleton maps of input images and captures the compression representations of human semantic information. We first concatenate the features from \textit{Res-2} to \textit{Res-5} by upsampling to the same resolution. Then the channel shuffle operation \cite{su2019multi} is performed to enhance channel interdependence from different levels. After that, the shuffled features are split and downsampled to their original resolutions. At last, we introduce a bottom-up structure to generate the skeleton heat maps and local compression features. 

As mentioned above, the skeleton is a kind of representation of the structural information. In fact, other structural features are also efficient in our experiments, such as the center of human parts. Furthermore, the thought of structural compression and fusion can also be applied to other datasets. In general, the cost of skeleton annotation is much lower than that for segmentation. 

\textbf{Boundary Decoder Branch.} This module aims at obtaining boundary-maps and the global boundary-aware feature representations to further improve the basic parsing prediction. As shown in Fig.~\ref{fig:overall structure}, the features of \textit{Res-3}, \textit{Res-4} are first upsampled to the same resolution as \textit{Res-2}. The feature maps are then concatenated together after the channel reduced operation. We use a sequential $1\times 1$ convolutional layers for the concatenated feature to better decode the extracted features. Finally, we get a 2-channel score map as the prediction of the semantic boundary.

\textbf{Compression and Expansion Group.}
The LCM aims at manipulating each part to a reasonable location, while the GEM aims at obtaining more clean boundaries. In order to construct a complete parsing system utilizing global and location information, we fuse features from LCM and GEM. Specifically, we concatenate the output features of LCM and GEM, then feed them to two $1\times 1$ convolutional layers to accomplish feature fusion. At last, a convolutional layer is deployed to get final parsing prediction. 

\textbf{Loss Function.} Our model is learned in an end-to-end manner. There are four outputs of the network: boundary prediction, skeleton prediction, coarse and fine parsing results. In general, the loss function of our model can be formulated as:
\begin{align}
L = L_{par\_base} + L_{par\_fine} +\alpha{L_{bd}}+\beta{L_{ske}},\;
\end{align}
where $L_{bd}$ denotes the weighted cross entropy loss function between boundary map from boundary decoder branch; $L_{ske}$ denotes the L2 loss for skeleton heat map from skeleton extraction branch; $L_{par\_base}$ denotes the cross entropy loss function for basic parsing results from basic parsing network; And $L_{par\_fine}$ denotes the cross entropy loss function with online hard example mining strategy proposed in \cite{shrivastava2016training} for the final parsing results, which is improved by enhanced attention group. $\alpha$ and $\beta$ are hyper-parameters used to balance weights of losses. 

\section{Experiments}

We demonstrate the effectiveness of our methods on two challenging benchmarks, i.e, LIP \cite{gong2017look} (single person parsing dataset) and Pascal-Person-Part \cite{chen2014detect} (multiple people parsing dataset).

\textbf{LIP} has a total of 50462 fine-grained images, within which a person is split to 19 semantic parts (e.g.\ , hair, face, left-/right-arms, left-/right-legs, scarf, jumpsuits, etc.). Besides, as a pose estimation dataset, LIP also contains human skeleton joint data (e.g.\ , head, neck, left-/right-shoulder, etc.). The images in LIP are collected from the real-world scenarios, which contain human appearing with challenging poses and views, heavily occlusions, various appearances, and low-resolutions. The whole dataset is divided into training, validation and testing phase, which contains 30,462, 10000 and 10000 images, respectively.

\textbf{Pascal-Person-Part} includes 3533 human images from PASCAL VOC 2010 \cite{everingham2010pascal}, which has large variation of human poses, scales, and occlusion. This dataset contains 14 joint and 7 part annotations. According to the standard split of this dataset, there are 1,716 images in the training set and the left 1817 images are in the testing set.

\begin{figure}[t]
	\begin{center}
		\begin{tabular}{cc}
			\hspace{-1ex}
			\vspace{-1.6ex}
			\subfigure{
				\includegraphics[width=0.47\textwidth, height=1.0in]{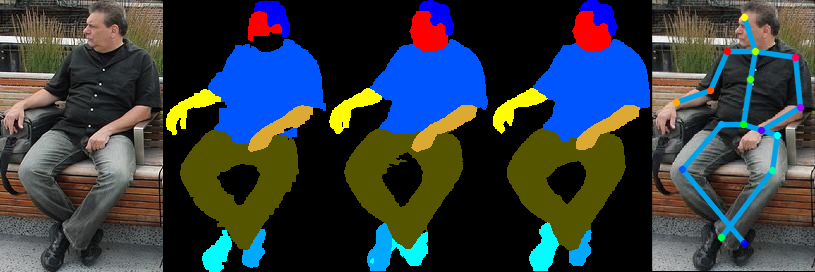}
			}  &
			\hspace{1ex}
			\subfigure{
				\includegraphics[width=0.47\textwidth, height=1.0in]{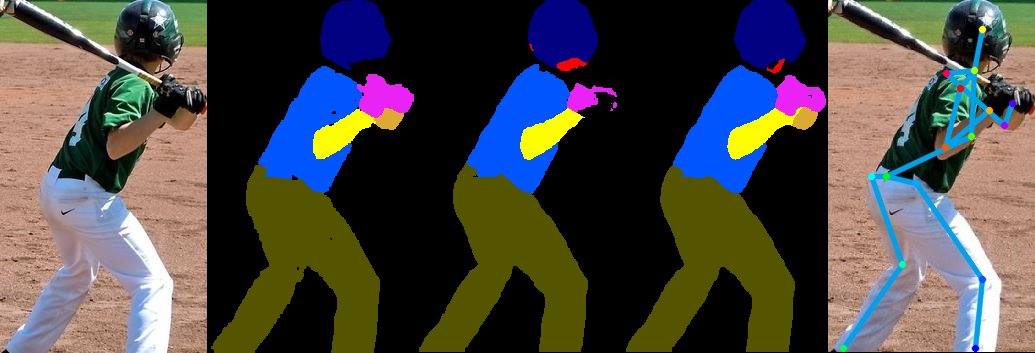}
			} \\	
		\end{tabular}		
	\end{center}
	\vspace{-4mm}
	\caption{Examples of the behavior of LCM. In each group, input image, ground truth, baseline result, our result, and skeleton prediction of our model is visualized. }
	\vspace{-0.5mm}
	\label{fig:lcm_fig}
\end{figure}

\begin{table}[t]
	\centering
	\caption{Detailed performance comparison of Local Compression Module. ``GC" means global convolutional block. Models are evaluated on LIP \texttt{validation} set.}
	\label{table:lcm_com}
	\setlength{\tabcolsep}{15 pt}
	\vspace{-2.5mm}
	\begin{tabular}{c|c|c}
		\hline
		Method & input size & mIoU \\
		\hline
		\hline
		Basic Network & $384\times 384$ &53.8\\
		Self-attention\cite{fu2019dual} (channel) &$384\times 384$& 55.2\\
		Concatenate (parsing \& skeleton) &$384\times 384$& 55.4 \\
		Attention (SENet \cite{hu2018squeeze}, our impl.) &$384\times 384$& 55.8\\
		\hline
		\hline
		LCM (w/o GC, $N$=16)&$384\times 384$& 56.1\\
		LCM (w/ GC, $N$=16)&$384\times 384$& 56.4\\
		LCM (w/ GC, $N$=32)&$384\times 384$& 56.6\\
		LCM (w/ GC, $N$=64)&$384\times 384$& \textbf{56.9}\\
		\hline
	\end{tabular}
	\vspace{-1mm}
\end{table}

\subsection{Implementation Details}

We use the ResNet-101~\cite{he2016deep} as backbone for feature extraction, initialized with weights learned on ImageNet~\cite{deng2009imagenet}. We train the network using mini-batch stochastic gradient descent (SGD) with batch size 40, momentum 0.9 and weight decay 0.0005. The base learning rate is set to $0.007$. We first train the network with a linear increasing warm-up strategy, and then use the ``poly" iter learning rate policy where the base rate is multiplied by $(1 - \frac{current_{iters}}{total_{iters}})^{power}$, in which $power=0.9$. For data augmentation, we adopt randomly left-right flipping and random scaling between 0.5 and 2 for all datasets during training. In evaluation, we apply the multi-scale inputs and also horizontally flip them.

For measuring the performance of our proposed network, we apply the mean pixel intersection-over-union (mIoU) as the metric. Besides, we also use pixel-wise mean accuracy (acc.) and part-wise mean accuracy (mean acc.) as additional metrics when compared with previous methods. 

\subsection{Ablation Study}
Here we perform comprehensive studies on LIP to reveal the effectiveness of each proposed module. We use the same training policy as mentioned above in the following experiments. For fair comparisons, the input size of experiments in this section is set to $384\times 384$.

\begin{figure}[t]
	\begin{center}
		\begin{tabular}{cc}
			\hspace{-1ex}
			\vspace{-1.6ex}
			\subfigure{
				\includegraphics[width=0.47\textwidth, height=1.0in]{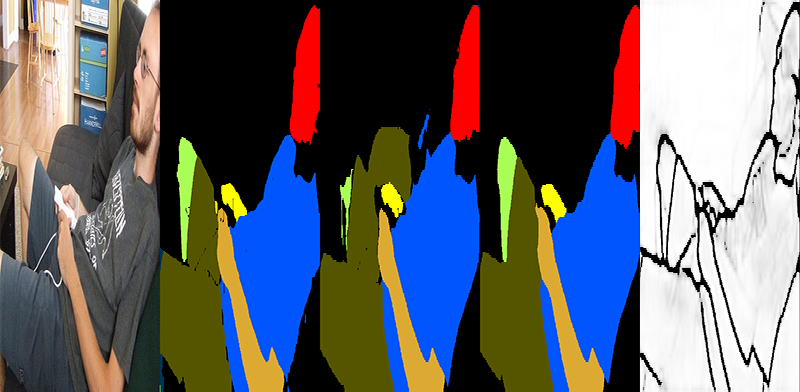}		
			}  &
			\hspace{1ex}
			\subfigure{
				\includegraphics[width=0.47\textwidth, height=1.0in]{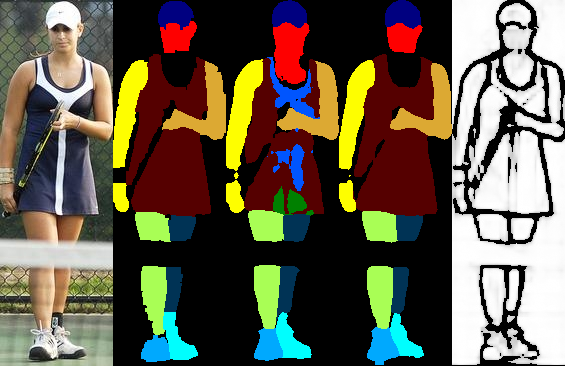}
			} \\	
		\end{tabular}
	\end{center}
	\vspace{-4mm}
	\caption{Examples of the behavior of GEM. In each group, input image, ground truth, baseline result, our result, and boundary prediction of our model is visualized. }
	\label{fig:gem_fig}
\end{figure}

\begin{table}[t]
	\centering
	\caption{Detailed performance comparison of our proposed Global Expansion Module. Models are evaluated on LIP \texttt{validation} set.}
	\label{table:gem_com}
	\setlength{\tabcolsep}{15 pt}
	\begin{tabular}{c|c|c}
		\hline
		Method & input size & mIoU \\
		\hline
		\hline
		Basic Network & $384\times 384$&53.8\\
		Self-attention~\cite{fu2019dual} (spatial) &$384\times 384$& 55.4\\
		Concatenate (parsing \& boundary) &$384\times 384$& 55.1\\
		\hline
		\hline
		GEM ($D$=16)&$384\times 384$& 55.6\\
		GEM ($D$=32)&$384\times 384$& 55.9\\
		GEM ($D$=64)&$384\times 384$& \textbf{56.4}\\
		\hline
	\end{tabular}
	\vspace{-1mm}
\end{table}

\subsubsection{Local Compression Module}
\label{sec_lcm}

As mentioned above, we use ResNet-101 followed by a Pyramid Pooling Module (PPM) as our baseline. Besides, we implement the channel-wise self-attention module proposed in \cite{fu2019dual} on the parsing feature, as a competitor of our LCM. Similarly, we implement the SENet proposed in~\cite{hu2018squeeze}, which has been widely used as a powerful attention module. As shown in Tab.~\ref{table:lcm_com}, LCM outperforms these attention modules with a certain margin, which verifies that the local compression guide from the skeleton domain is more powerful than attention from the parsing domain.

We also concatenate parsing and skeleton features directly. From Tab.~\ref{table:lcm_com}, adding skeleton features can improve parsing accuracy by 1.4\%, but still is 1.5\% lower than our LCM. This indicates that our proposed LCM is a better local information aggregation module than direct combination, and the improvement of LCM is much better than simply adding extra features to the baseline.

We further tune some hyper-parameters in LCM. We found that add a global convolutional layer can improve the result by 0.3\%. Based on these settings, we further add $N$ from 16 to 32 and 64, achieving 56.6\% and 56.9\% respectively. Since $N$ denotes the strength of the local compression module, this result shows that increasing the power of LCM can improve parsing results. Due to the limitation of GPU memory and in order to keep a relatively favorable batch size, we don't further increase $N$ and use $N=64$ in later experiments.

In Fig.\ref{fig:lcm_fig}, we visualize two groups of validation images to show the help of LCM to parsing. In each group, images are input, ground truth, the output of the baseline, the output of our model, and skeleton estimation overlaid on the input image. In the first group, we can see that with the help of the right skeleton estimation, our model can parse the left foot and right foot correctly. In the second group, the baseline method can not parse hands because of occlusion, while our method obtain correct result with the help of local compression module.

\begin{table}[t]
	\centering
	\caption{Ablation study (LCM \& GEM). Models are tested on LIP \texttt{validation} set. }
	\vspace{-1.5mm}
	\setlength{\tabcolsep}{12 pt}
	\begin{tabular}{c|c|cc|c}
		\hline
		Backbone & input size & LCM & GEM & mIoU \\
		\hline
		\hline
		ResNet-101 &$384\times 384$& $\times$ & $\times$ & 53.8\\
		ResNet-101 &$384\times 384$& $\checkmark$ & $\times$ & 56.4\\
		ResNet-101 &$384\times 384$& $\times$ & $\checkmark$ & 56.9\\
		ResNet-101 &$384\times 384$& $\checkmark$ & $\checkmark$ & \textbf{57.7}\\
		\hline
	\end{tabular}
	\vspace{-0mm}
	\label{table:ablation}
\end{table}

\begin{table}[t]
	\small
	\setlength{\tabcolsep}{2pt}
	\centering
	\caption{Comparison with state-of-the-art methods on the \texttt{validation} set of LIP.}
	\vspace{-1mm}
	\begin{tabular}{c|c|c|c|c}
		\hline
		Method & Input size & pixel acc. & mean acc. & mIoU \\	
		\hline
		DeepLab (VGG-16) \cite{deeplab} & $384\times 384$ & 82.7 & 51.6 & 41.6 \\
		DeepLab (ResNet-101) \cite{deeplab} & $384\times 384$ & 84.1 & 55.6 & 44.8 \\ 
		JPPNet \cite{liang2018look} & $384\times 384$ & 86.4 & 62.3 & 51.1 \\
		CE2P \cite{ruan2019devil} & $473\times 473$ & 87.4 & 63.2 & 53.1 \\
		HRNet \cite{sun2019high} & $473\times 473$ & 88.2 & 67.4 & 55.9 \\
		\hline
		Ours & $384\times 384$ & \textbf{88.6} & \textbf{68.6} & \textbf{57.7} \\
		Ours & $473\times 473$ & \textbf{88.9} & \textbf{68.8} & \textbf{58.1} \\
		\hline
	\end{tabular}
	\vspace{-1mm}
	\label{table:lip_sota}
\end{table}

\subsubsection{Global Expansion Module}
\label{sec_gem}

We use the same basic network as in experiments above. We implementation spatial self-attention from \cite{fu2019dual} on parsing feature, which learns a spatial attention map using parsing feature and multiplies it to parsing feature. As shown in Tab.~\ref{table:gem_com}, our GEM also outperforms this module, proving that boundary guided attention is more helpful for human parsing. 

Combining parsing feature and boundary feature directly is less useful than GEM, verifying that GEM is a more effective way to exploit information from global boundaries. This result shows that the attentional boundary features is not the key to the improvement of GEM; On the contrary, it is more important to find the effective way to fully utilize the boundary information in GEM.

As the measure of the strength of the global expansion force, the value of parameter $D$ can strongly affect the final result of parsing, As shown in Tab.~\ref{table:gem_com}. We study the result of changing the value of $D$ from 16 to 64. Not surprisingly, set $D$ to 64 achieves better accuracy. Again, we don't further increase $D$ due to the hardware limitation and use $D=64$ in later experiments by default. 

To validate the effectiveness of the proposed GEM, we show some results in Fig.~\ref{fig:gem_fig}. In each group, images are input, ground truth, the output of the baseline, the output of our model, and the learned boundaries. In the first group, the baseline method cannot find the right boundary of the right leg. With the expansive force from the proposed GEM, our parsing system can find the right boundaries between the right leg and background, leading to a more complete result. In the second group, our model can improve inner consistency in clothes, by removing noise in a part.

\subsubsection{Module Fusion with Networks}

Here we conduct experiments on fusion LCM and GEM into an end-to-end network. As shown in Tab.~\ref{table:ablation}, Aggregating LCM and GEM can further boost parsing accuracy. Since LCM can be treated as a channel-wise aggregator, while GEM is a spatial separator, they are complementary and can improve parsing from different aspects. 

\begin{table}[t]
	
	\caption{Per-class comparison of mIoU with state-of-the-art methods on the \texttt{validation} set of LIP.}
	\tiny
	\renewcommand\arraystretch{1.05}
	\setlength{\tabcolsep}{1.5 pt}
	\centering
	
	\begin{tabular}{c|ccccccccccc|cc}
		\hline
		
		\multirow{2}{*}{Class} & SegNet & FCN-8s  & Attention  & SSL  & DeepLab  & ASN  & SSL  & MMAN  & SS-NAN  & JPPNet  & CE2P \ & ours  & ours \\
		&  \cite{badrinarayanan2017segnet} &\cite{long2015fully}&\cite{chen2016attention}&\cite{gong2017look}&\cite{chen2017deeplab}&\cite{luc2016semantic}&\cite{gong2017look}& \cite{luo2018macro} &\cite{zhao2017self}&\cite{liang2018look}&\cite{ruan2019devil}& 384x384&473x473\\
		\hline
		\hline
		hat& 26.60&39.79 &58.87 &59.75 &59.76 &56.92 &58.21 &57.66 &63.86 &63.55 &65.29 &69.45 &\textbf{70.24} \\
		hair& 44.01&58.96 &66.78 &67.25 &66.22 &64.34 &67.17 &65.63 &70.12 &70.20 &72.54 &74.03 &\textbf{74.46} \\
		glove&0.01 &5.32 &23.32 &28.95 &28.76 &28.07 &31.20 &30.07 &30.63 &36.16 &39.09 &44.15 &\textbf{44.22} \\
		s-glass& 0.00& 0.08& 19.48& 21.57& 23.91& 17.78& 23.65& 20.02& 23.92& 23.48& 32.73&\textbf{36.96} &36.06 \\
		u-clot& 34.46& 49.08& 63.20& 65.30& 64.95& 64.90& 63.66& 64.15& 70.27& 68.15& 69.46& 71.93&\textbf{72.26} \\
		dress& 0.00& 12.36& 29.63& 29.49& 33.68& 30.85& 28.31& 28.39& 33.51& 31.42& 32.52& 38.60&\textbf{39.58} \\
		coat& 15.97& 26.82& 49.70& 51.92& 52.86& 51.90& 52.35& 51.98& 56.75& 55.65& 56.28& 59.14&\textbf{59.81} \\
		sock& 3.59& 15.66& 35.23& 38.52& 37.67& 39.75& 39.58& 41.46& 40.18& 44.56& 49.67& 52.58&\textbf{53.39}\\
		pant& 33.56& 49.41& 66.04& 68.02& 68.05& 71.78& 69.40& 71.03& 72.19& 72.19& 74.11& 76.67& \textbf{77.02}\\
		j-suit& 0.01& 6.48& 24.73& 24.48& 26.15& 25.57& 28.61& 23.61& 27.68& 28.39& 27.23& 31.91& \textbf{31.99}\\
		scarf& 0.00& 0.00& 12.84& 14.92& 17.44& 7.97& 13.70& 9.65& 16.98& 18.76& 14.19& 27.59& \textbf{27.82}\\
		skirt& 0.00& 2.16& 20.41& 24.32& 25.23& 17.63& 22.52& 23.20& 26.41& 25.14& 22.51&\textbf{31.44} &31.27 \\
		face& 52.38& 62.65& 70.58& 71.01& 70.00& 70.77& 74.84& 69.54& 75.33& 73.36& 75.50& 76.63& \textbf{76.74}\\
		l-arm& 15.30& 29.78& 50.17& 52.64& 50.42& 53.53& 52.83& 55.30& 55.24& 61.97& 65.14& 69.94& \textbf{70.32}\\
		r-arm& 24.23& 36.63& 54.03& 55.79& 53.89& 56.70& 55.67 &58.13 &58.93 &63.88 &66.59 &71.68& \textbf{72.12}\\
		l-leg& 13.82& 28.12& 38.35& 40.23& 39.36& 49.58& 48.22& 51.90& 44.01& 58.21& 60.10& 64.67& \textbf{65.52}\\
		r-leg&13.17&26.05&37.70&38.80&38.27&48.21&47.49&52.17&41.87&57.99&58.59&63.89&\textbf{64.08}\\
		l-shoe& 9.26& 17.76& 26.20& 28.08& 26.95& 34.57& 31.80& 38.58& 29.15& 44.02& 46.63& 51.88& \textbf{52.64}\\
		r-shoe& 6.47& 17.70& 27.09& 29.03& 28.36& 33.31& 29.97& 39.05& 32.64& 44.09& 46.12& 52.28& \textbf{52.88}\\
		bkg& 70.62& 78.02& 84.00& 84.56& 84.09& 84.01& 84.64& 84.75& 88.67& 86.26& 87.67& 88.87& \textbf{88.96}\\
		\hline
		\hline
		mIoU& 18.17& 28.29& 42.92& 44.73& 44.80& 45.41& 46.19& 46.81& 47.92& 51.37& 53.10& 57.72& \textbf{58.07}\\		
		\hline
	\end{tabular}
	\vspace{-1mm}
	\label{table:1}
\end{table}

\subsection{Results on LIP Dataset}

\textbf{Comparing with State-of-the-Arts.} Here we compare the ACENet with other state-of-the-art methods. We use hyper-parameters learned from ablation studies and use input size with $384\times 384$ and $473\times 473$ to fairly compare with all methods. We show methods with detailed per-class IoU in Tab.~\ref{table:1}, and methods with pixel accuracy and mean accuracy in Tab.~\ref{table:lip_sota}. From Tab.~\ref{table:1}, we can see that our method improves visibly on parts like gloves, dress, coat, scarf, legs, and arms, validating the effectiveness of skeleton-based LCM and boundary-based GEM. From Tab.~\ref{table:lip_sota}, we can see that our method with input size set to $384\times 384$ outperforms other compared methods. With input size $473\times 473$, our method achieves a new state-of-the-art mIoU on LIP.

\begin{figure*}[t]
	{\tiny \hspace{1.5em}\textbf{Input}\hspace{2em}\textbf{Label} \hspace{1.2em}\textbf{Baseline}\hspace{1em}\textbf{CE2P}\cite{ruan2019devil} \hspace{0.3em} \textbf{Ours} \hspace{3.5em}\textbf{Input}\hspace{2em}\textbf{Label} \hspace{1.2em}\textbf{Baseline}\hspace{1em}\textbf{CE2P}\cite{ruan2019devil} \hspace{0.5em} \textbf{Ours}} 
	\vspace{-3mm}
	
	\begin{center}
		\begin{tabular}{cc}
			\hspace{-2.2ex}
			\subfigure{
				\begin{tabular}{l}
					\includegraphics[width=.46\textwidth, height= 0.6in]{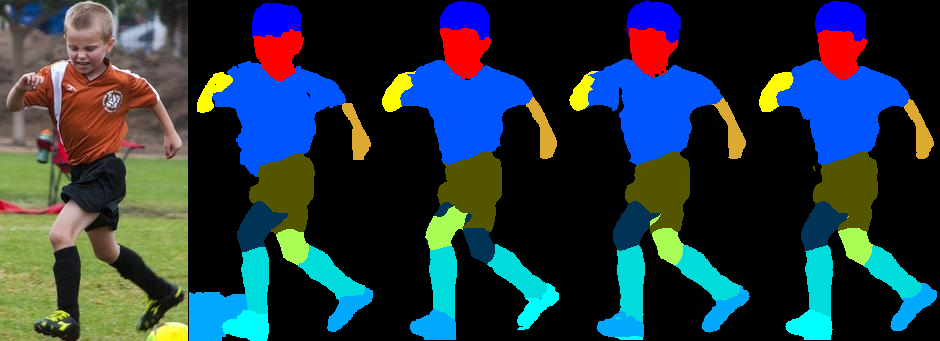}
				\end{tabular}		
			} \vspace{-2.5ex} &
			\subfigure{
				\begin{tabular}{r}			
					\includegraphics[width=.46\textwidth, height= 0.6in]{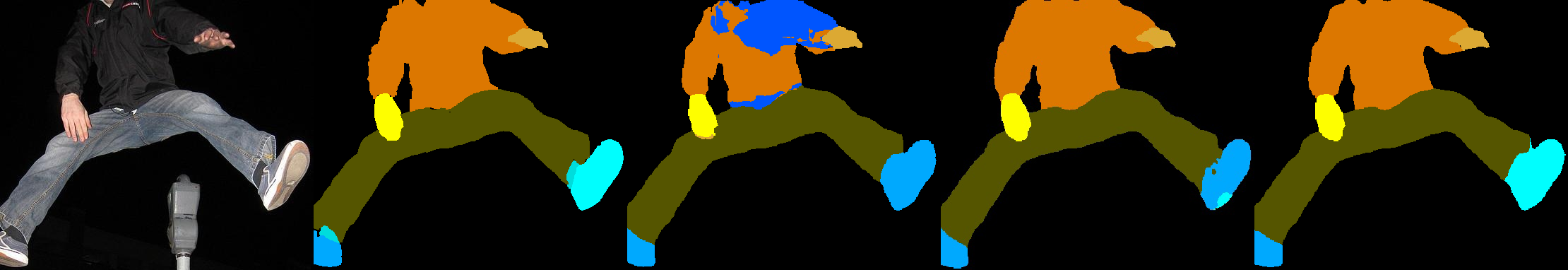}
				\end{tabular}
			} \\
			\vspace{-0.2mm}
			{\scriptsize (a)} 
			\vspace{-0.0ex}&\hspace{-4ex}{\scriptsize (b)}\\
			
			\hspace{-2.2ex}
			\subfigure{
				\begin{tabular}{l}					
					\includegraphics[width=.46\textwidth, height= 0.6in]{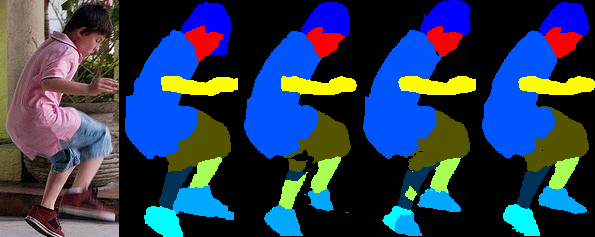}
				\end{tabular}
			} \vspace{-2.5ex}&
			\subfigure{
				\begin{tabular}{r}					
					\includegraphics[width=.46\textwidth, height= 0.6in]{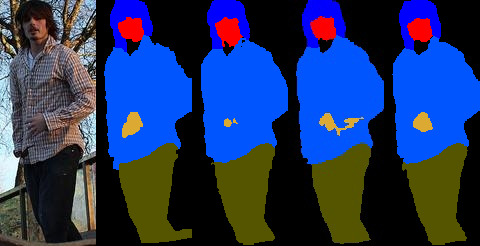}
				\end{tabular}
			} \\
			\vspace{-0.2mm}
			{\scriptsize (c)} 
			\vspace{-0.0ex}&\hspace{-4ex}{\scriptsize (d)}\\
			
			\hspace{-2.2ex}
			\subfigure{
				\begin{tabular}{l}					
					\includegraphics[width=.46\textwidth, height= 0.6in]{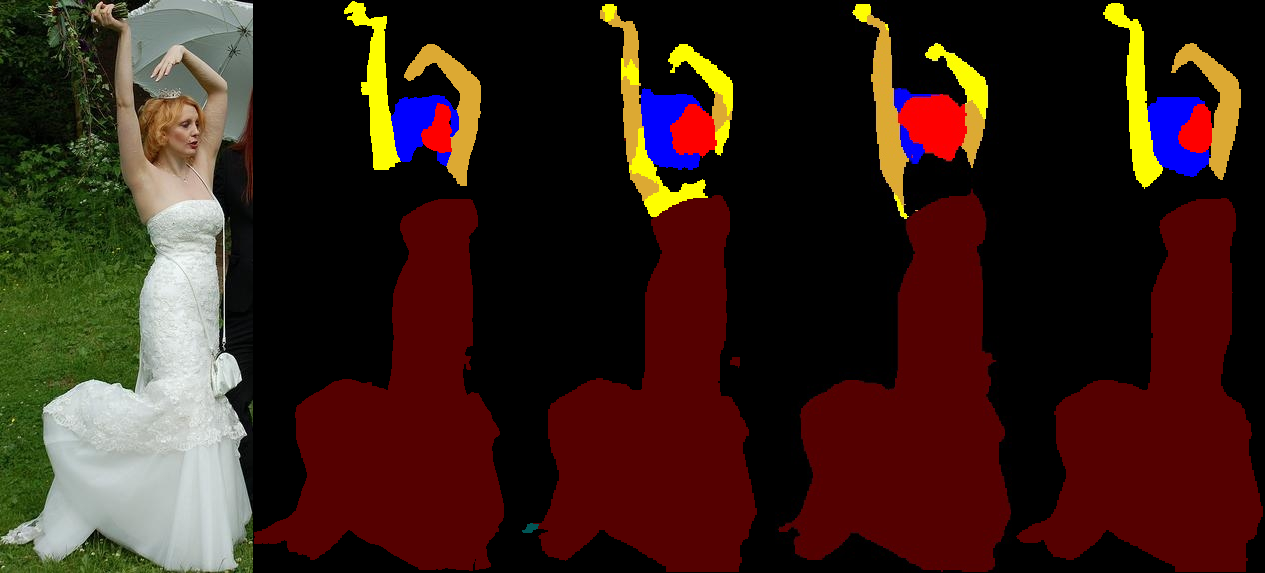}
				\end{tabular}
			}  \vspace{-2.5ex}&
			\subfigure{
				\begin{tabular}{r}					
					\includegraphics[width=.46\textwidth, height= 0.6in]{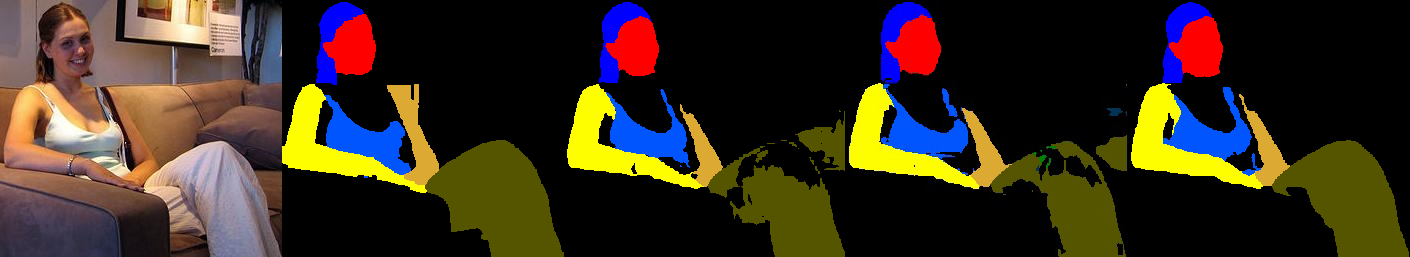}
				\end{tabular}
			} \\
			\vspace{-0.5mm}
			{\scriptsize (e)} 
			\vspace{-0.0ex}&\hspace{-4ex}{\scriptsize (f)}\\
			
			\hspace{-2.2ex}
			\subfigure{
				\begin{tabular}{l}					
					\includegraphics[width=.46\textwidth, height= 0.6in]{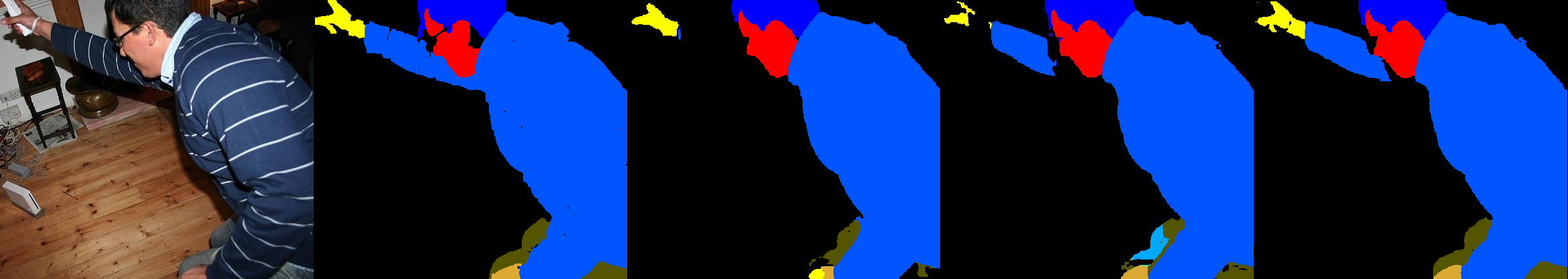}
				\end{tabular}
			}  \vspace{-2.5ex}&
			\subfigure{
				\begin{tabular}{r}					
					\includegraphics[width=.46\textwidth, height= 0.6in]{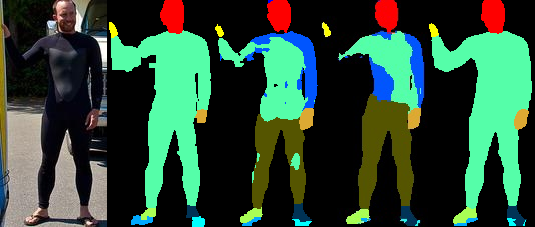}
				\end{tabular}
			} \\
			\vspace{-2mm}
			{\scriptsize (g)} 
			\vspace{-0.0ex}&\hspace{-4ex}{\scriptsize (h)}\\
			
		\end{tabular}
	\end{center}
	\begin{center}
		\vspace{-1.5mm}
		\includegraphics[width=.95\textwidth, height=0.15in]{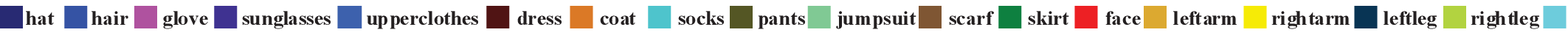}
	\end{center}
	\vspace{-5.5mm}	
	\caption{Qualitative Comparison on LIP \texttt{validation} dataset. }
	\vspace{-0mm}
	\label{fig:q_com}
\end{figure*}

\textbf{Qualitative Comparison.} As shown in Fig.~\ref{fig:q_com}, we visualize the outputs of ACENet, and qualitatively compare the model with other methods on LIP. In each subfigure, images are input image, label image, baseline result, CE2P~\cite{ruan2019devil} result, and our result respectively. In (a), the baseline and CE2P cannot parse left-hand and upper-clothes while our model works much better. In (b), the compared methods cannot predict a clean boundary between the coat and other parts. In (c), other methods are failed to learn a complete left-leg. In (d), baseline and CE2P are failed to recognize the left-arm due to similar color with upper-clothes. In (e), our model can predict the left-arm and right-arm better, with the help of local compression module. In (f), supervision from the global expansion module makes the parsing system find the right boundary between right-leg and background. In (g), the prediction of legs, upper-clothes, and right-arm is well improved. In (h), our model can learn an integrated jumpsuit, which is much better than others. Besides, in (a) and (c), misclassifications of left and right shoes are largely alleviated. ACENet can solve these bad cases by using local compression and global expansion guidance, leading to a better parsing scheme.

\subsection{Results on Pascal-Person-Part Dataset}
To show the generality of the proposed ACENet, we also report the results on the Pascal-Person-Part dataset in Table~\ref{table:ppp}. Based on the best ACENet architecture studied above, we evaluate experiments on the Pascal-Person-Part benchmark. We use the settings mentioned above for Pascal-Person-Part. During training and testing, we use input images and skeleton data from the original dataset, while the boundary data generated from the ground truth. We adopt the policy used in \cite{ruan2019devil}, employing a Mask R-CNN \cite{he2017mask} to extract all the person patches in input images. Each patch, as well as its corresponding skeleton patches, are fed to the ACENet as input. As shown in Tab~\ref{table:ppp}, ACENet can achieve 70.9 mIoU on Pascal-Person-Part, which is superior to almost all the other state-of-the-art methods. Although the mean IoU in~\cite{gong2019graphonomy} is a little higher, it adopts a more complex base model while we only adopt ResNet-101, and uses extra images in CIHP dataset~\cite{gong2018instance}. Therefore, the experimental results can demonstrate the generalization ability of ACENet, and the proposed method can handle complex images with a large variety of human poses, scales, and occlusion.

\begin{table}[t]
	\centering
	\setlength{\tabcolsep}{18pt}
	\caption{Comparison with state-of-the-art methods on \texttt{test} set of Pascal-Person-Part.}
	\begin{tabular}{c|c|c}
		\hline
		Method & Backbone & mIoU \\
		
		\hline
		Attention \cite{chen2016attention} & VGG-16 & 56.39 \\
		HAZN \cite{xia2016zoom} & VGG-16 & 57.54 \\
		LG-LSTM \cite{liang2016semantic} & VGG-16 & 57.97 \\
		Attention+SSL\cite{gong2017look} & ResNet-101 & 59.36 \\
		Attention+MMAN \cite{luo2018macro} & VGG-16 & 59.91 \\
		G-LSTM \cite{liang2016semantic2} & VGG-16 & 60.16 \\
		SS-NAN \cite{zhao2017self} & ResNet-101 & 62.44 \\
		Weakly Supervised \cite{yang2019weakly} & DeepLab & 62.05 \\
		Structure LSTM \cite{liang2017interpretable} & VGG-16 & 63.57 \\
		JPS \cite{xia2017joint} & ResNet-101 & 64.39 \\
		MuLA \cite{nie2018mutual} & Hourglass & 65.10 \\
		PGN \cite{gong2018instance} & DeepLab-v2 & 68.40 \\
		Graphonomy \cite{gong2019graphonomy} & Xception & 71.14 \\
		\hline
		Ours & ResNet-101 & \textbf{70.86} \\
		\hline
	\end{tabular}
	
	\label{table:ppp}
\end{table}

\section{Conclusion}
In this paper, we propose Local Compression Module (LCM) and Global Expansion Module (GEM) to improve human parsing performance by getting rid of ``intra-part inconsistency" and ``inter-part indistinction". We rethink the human parsing task by considering the characteristic structure and semantics among human parts. In this way, the LCM can make predicted parts to more reasonable locations through structural joint points, while the GEM can further improve the consistency in each part by the explicit supervision of semantic boundary. The ablation studies and visualizations of parsing results demonstrate that the proposed ACENet can obtain more accurate human parsing results with smooth parts and clear boundaries. Based on the proposed method, we achieve state-of-the-art performance on two challenging human parsing benchmarks. 

%
%
\bibliographystyle{splncs04}
\bibliography{humanparsing}
\end{document}